\documentclass[runningheads]{llncs}

\usepackage{eccv}

\usepackage{eccvabbrv}

\usepackage{graphicx}
\usepackage{booktabs}

\usepackage[accsupp]{axessibility}  

\usepackage[pagebackref,breaklinks,colorlinks,citecolor=eccvblue]{hyperref}

\usepackage{orcidlink}

\begin{document}

\title{A Tutorial on ALOS2 SAR Utilization: \\Dataset Preparation, Self-Supervised Pretraining, \\and Semantic Segmentation}

\titlerunning{A Tutorial on ALOS2 SAR Utilization}

\author{Nevrez Imamoglu\inst{1}$^{,*}$\orcidlink{0000-0002-2661-599X} \and
Ali Caglayan\inst{1}$^{,}$\thanks{Authors equally contributed to this work.}\orcidlink{0000-0002-3408-8659} \and
Toru Kouyama\inst{1}\orcidlink{0000-0002-1060-3986}}

\authorrunning{N.~Imamoglu \& A.~Caglayan  et al.}

\institute{National Institute of Advanced Industrial Science and Technology 
\email{\{nevrez.imgmoalu,ali.caglayan,t.kouyama@aist.go.jp\}@aist.go.jp}}

\maketitle

\begin{abstract}
Masked auto-encoders (MAE) and related approaches have shown promise for satellite imagery, but their application to synthetic aperture radar (SAR) remains limited due to challenges in semantic labeling and high noise levels. Building on our prior work with SAR-W-MixMAE, which adds SAR-specific intensity-weighted loss to standard MixMAE for pretraining, we also introduce SAR-W-SimMIM—a weighted variant of SimMIM applied to ALOS-2 single-channel SAR imagery. This method aims to reduce the impact of speckle and extreme intensity values during self-supervised pretraining. We evaluate its effect on semantic segmentation compared to our previous trial with SAR-W-MixMAE and random initialization, observing notable improvements.

In addition, pretraining and fine-tuning models on satellite imagery pose unique challenges, particularly when developing region-specific models. Imbalanced land cover distributions such as dominant water, forest, or desert areas can introduce bias, affecting both pretraining and downstream tasks like land cover segmentation. To address this, we constructed a SAR dataset using ALOS-2 single-channel (HH polarization) imagery focused on the Japan region, marking the initial phase toward a national-scale foundation model. This dataset was used to pretrain a vision transformer-based autoencoder, with the resulting encoder fine-tuned for semantic segmentation using a task-specific decoder.

Initial results demonstrate significant performance improvements compared to training from scratch with random initialization. In summary, this work provides a guide to process and prepare ALOS2 observations to create dataset so that it can be taken advantage of self-supervised pretraining of models and fine tuning downstream tasks such as semantic segmentation.

\end{abstract}

\section{Introduction}
\label{sec:intro}

Approaches such as masked auto-encoders (MAE) \cite{he2022masked} and their derivatives \cite{xie2022simmim,liu2023mixmae,hong2024spectralgpt,jakubik2023foundationmodelsgeneralistgeospatial} have demonstrated efficacy in applications to satellite imagery. In remote sensing, most ongoing technical validation of foundation models has focused on optical data, such as RGB or multi-spectral images. Synthetic aperture radar (SAR) data, however, remains less explored due to two primary challenges: (1) the difficulty of semantic labeling for dataset creation, and (2) the inherently higher noise content compared to optical imagery. In our previous work, we took advantage of MixMAE \cite{liu2023mixmae}, a variant of masked auto-encoder that mixes two images for masked regions, and extended it by leveraging SAR-specific physical characteristics to apply intensity-based weighting (SAR-W-MixMAE \cite{sar_w_mixmae_igarss2025,sar_w_mixmae_ieeejstars2026,rssj2025sarwmixmae,caglayan2026enhancedlulcsegmentationlightweight}) to the auto-encoder training loss (mean absolute error).

In this study, we extend this concept to a more standard masked auto-encoder variant, SimMIM \cite{xie2022simmim} (where masked regions are zeroed), applied to ALOS-2 single-channel SAR imagery. Our approach, SAR-W-SimMIM, aims to mitigate the impact of speckle and extreme intensity values in SAR images. The primary objective is to evaluate the effect of self-supervised pretraining with SimMIM on semantic segmentation of SAR data and compare results against SAR-W-MixMAE \cite{sar_w_mixmae_igarss2025,sar_w_mixmae_ieeejstars2026} and randomly initialized models. The SAR intensity-based weighting of the reconstruction loss has yielded encouraging results in both self-supervised SAR pretraining and downstream segmentation tasks.

In addition, pretraining and fine-tuning models on satellite imagery pose unique challenges, particularly when developing region-specific models. Imbalanced land cover distributions—such as dominant water, forest, or desert areas—can introduce bias, affecting both pretraining and downstream tasks like land cover segmentation. To address this, we constructed a SAR dataset using ALOS-2 single-channel (HH polarization) imagery focused on the Japan region, marking the initial phase toward a national-scale foundation model. This dataset was used to pretrain a vision transformer-based autoencoder, with the resulting encoder fine-tuned for semantic segmentation using a task-specific decoder. Initial results demonstrate significant performance improvements compared to training from scratch with random initialization.

Foundation model pretraining typically relies on large-scale image datasets, as seen in the development of various masked autoencoder models \cite{he2022masked}. However, in remote sensing, satellite observations often exhibit regional redundancy, both spatially and temporally, unless land use changes occur in the area \cite{jakubik2023foundationmodelsgeneralistgeospatial}. Moreover, random sampling can introduce bias in both the pretraining and fine-tuning stages, potentially affecting model performance and efficiency \cite{jakubik2023foundationmodelsgeneralistgeospatial,sar_w_mixmae_igarss2025,sar_w_mixmae_ieeejstars2026}.

In this work, we constructed a national-scale SAR foundation model pretraining dataset focused on the Japan region as in \cite{rssj2025sarwmixmae,caglayan2026enhancedlulcsegmentationlightweight}, using ALOS2 SAR imagery with HH polarization \cite{JAXA_LULC_2023}. The goal is to enable more efficient pretraining and fine-tuning for semantic segmentation tasks related to Japan's land use and land cover \cite{JAXA_LULC_2023}. Pretraining was conducted using the proposed SAR-W-SimMIM model similar to the previous SAR-W-MixMAE \cite{sar_w_mixmae_igarss2025,sar_w_mixmae_ieeejstars2026,rssj2025sarwmixmae,caglayan2026enhancedlulcsegmentationlightweight}. We initially evaluated downstream performance by comparing fine-tuning from the pretrained model against training from scratch with randomly initialized weights.

\section{Data Preparation}

\subsection{Prepraring SAR and Label patches}
To prepare ALOS2 Synthetic Aperture Radar (SAR) dataset for large-scale pretraining and semantic segmentation, a three steps of preprocessing is applied to extract image patches from the SAR observations \cite{rssj2025sarwmixmae,caglayan2026enhancedlulcsegmentationlightweight}. These steps includes downsampling, semantic label generation, and patch extraction from both SAR and label images for dataset construction (Source code will be available at: https://github.com/nevrez/ALOS2-SAR-data-processing-sampling). 

\begin{itemize}
    \item \textbf{Downsampling of SAR Imagery:} Original high-resolution SAR GeoTIFFs were first downsampled by half to reduce data size. Downsampling was performed by using average resampling of neighboring pixels while preserving the original metadata information such as geo-referencing information, coordinate reference system (CRS), etc. 
    \\
    \item \textbf{Generation of Semantic Labels from JAXA HR-LULC \cite{JAXA_LULC_2023}:} To obtain ground-truth label images for land cover classification, semantic labels were derived from the JAXA High-Resolution Land Use/Land Cover (HR-LULC) dataset \cite{JAXA_LULC_2023}. For each SAR scene, spatially matched with overlapping LULC tiles were merged, re-projected, and resampled to align precisely with the SAR image’s resolution and projection. Label rasters were clipped to the SAR coverage area, and regions corresponding to SAR no-data pixels were masked to ensure perfect pixel-wise correspondence. 
    \\
    \item \textbf{Patch-Based Dataset Generation:} The aligned SAR and semantic label GeoTIFFs were subdivided into non-overlapping and fixed-size patches (256 × 256 pixels) to create a dataset suitable for pretraining and finetuning tasks. Each patch pair maintains spatial alignment and geo-referencing metadata. To avoid the challenges on model trainings that can arise from no-data regions, patches containing no-data regions were excluded. SAR intensity values were also converted to the decibel (dB) scale using sensor-specific lookup tables to standardize the dynamic range across scenes. 
\end{itemize}


\begin{figure}[tb]
  \centering
  \includegraphics[height=5.5cm]{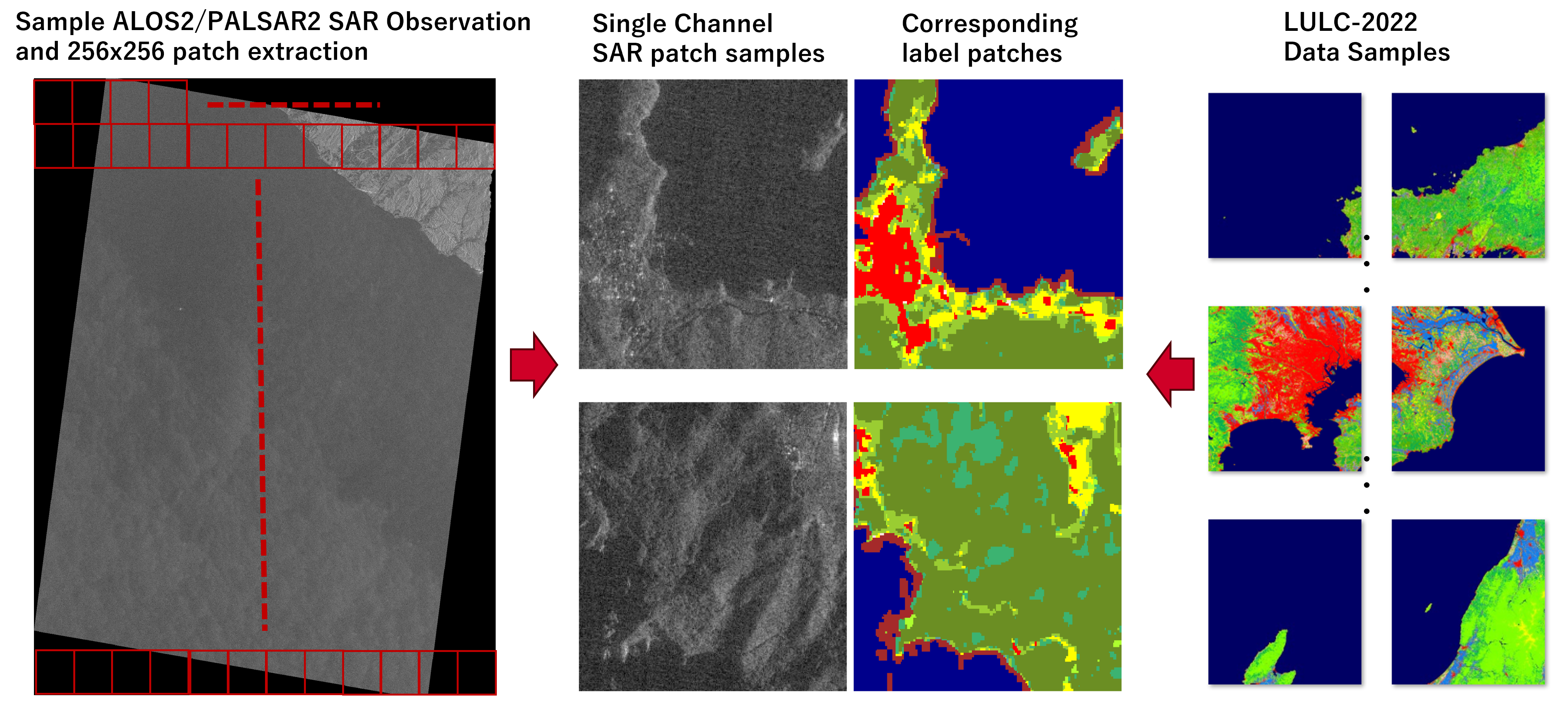}
  \caption{Selection of non-overlaping patches from the observations and obtaining corresponding JAXA-LULC \cite{JAXA_LULC_2023} label patches.  }
  \label{fig:patch_extraction}
\end{figure}

\begin{figure}[!b]
  \centering
  \includegraphics[height=5.75cm]{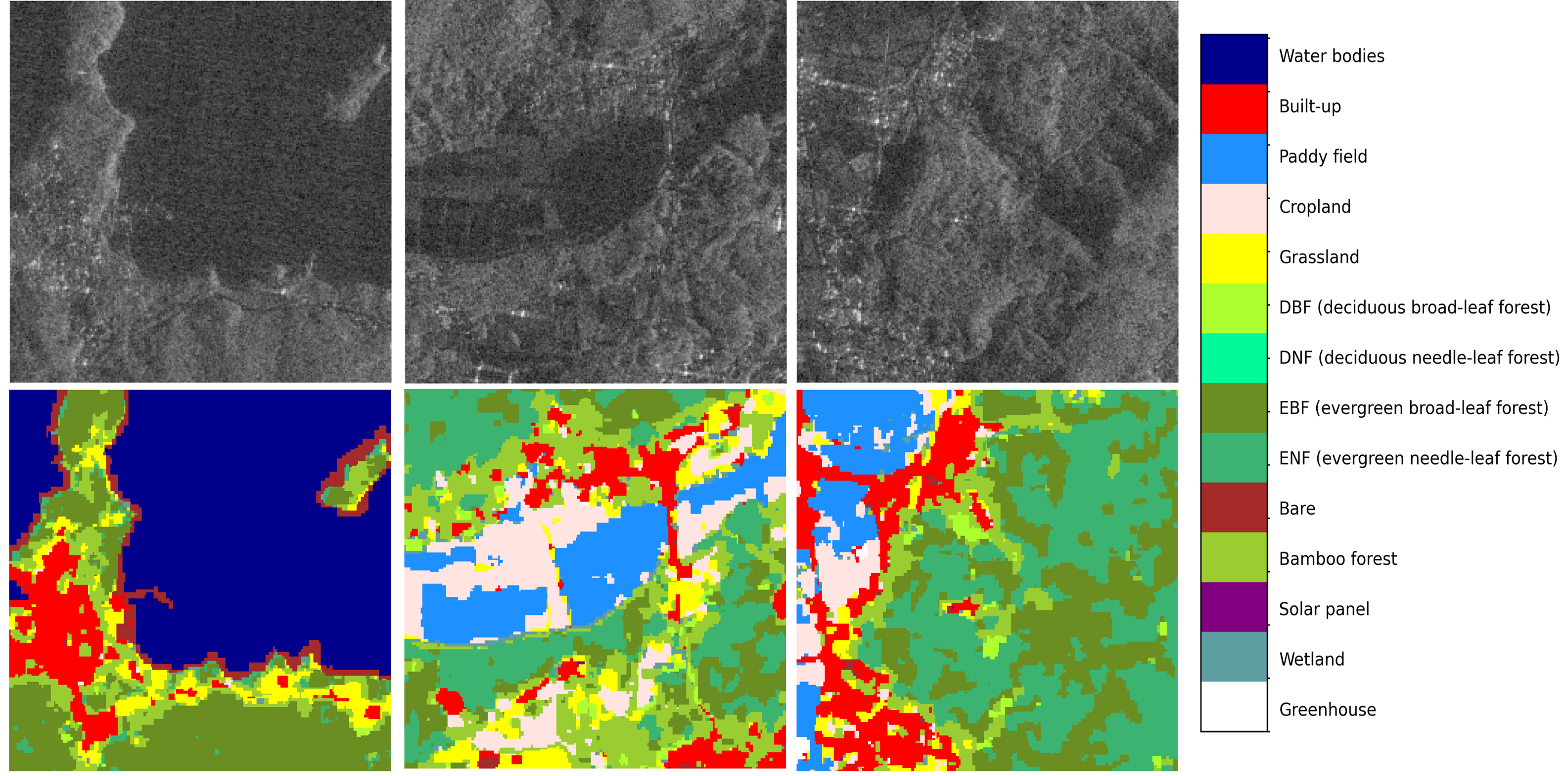}
  \caption{Sample SAR images and corresponding JAXA-LULC \cite{JAXA_LULC_2023} label patches.  }
  \label{fig:patch_samples}
\end{figure}

\subsection{Category aware data sampling}
\label{sec:category_sampling}
As also used in \cite{rssj2025sarwmixmae,caglayan2026enhancedlulcsegmentationlightweight}, we utilized a category distribution aware location sampling to improve representation learning performance and minimize class imbalance (Source code will be available at:
https://github.com/nevrez/ALOS2-SAR-data-processing-sampling). This is achieved by following steps:
\begin{itemize}
    \item \textbf{Calculate category weights:} 

    ~~~The total number of pixels is computed by summing all category counts from all of the JAXA-LULC \cite{JAXA_LULC_2023} data. Category distribution are then calculated as:
    \[
        P(\text{category}) = \frac{\text{pixel count for category}}{\text{total pixel count}}
    \]
    ~~~To tackle the category imbalance, weights are assigned inversely proportional to distribution of the categories:
    \[
    w(\text{category}) = \frac{1}{P(\text{category})}
    \]
    ~~~Thus, rarer categories receive higher weights. Finally, weights are normalized so that their sum equals 1:
    \[
    w_{\text{normalized}} = \frac{w(\text{category})}{\sum w(\text{all categories})}
    \]
    
    \item \textbf{Calculate number of samples for each LULC label image:}
    \item \textbf{Sample N number of locations based on category weights:}
\end{itemize}

\begin{figure}[!t]
  \centering
  \includegraphics[width=\textwidth]{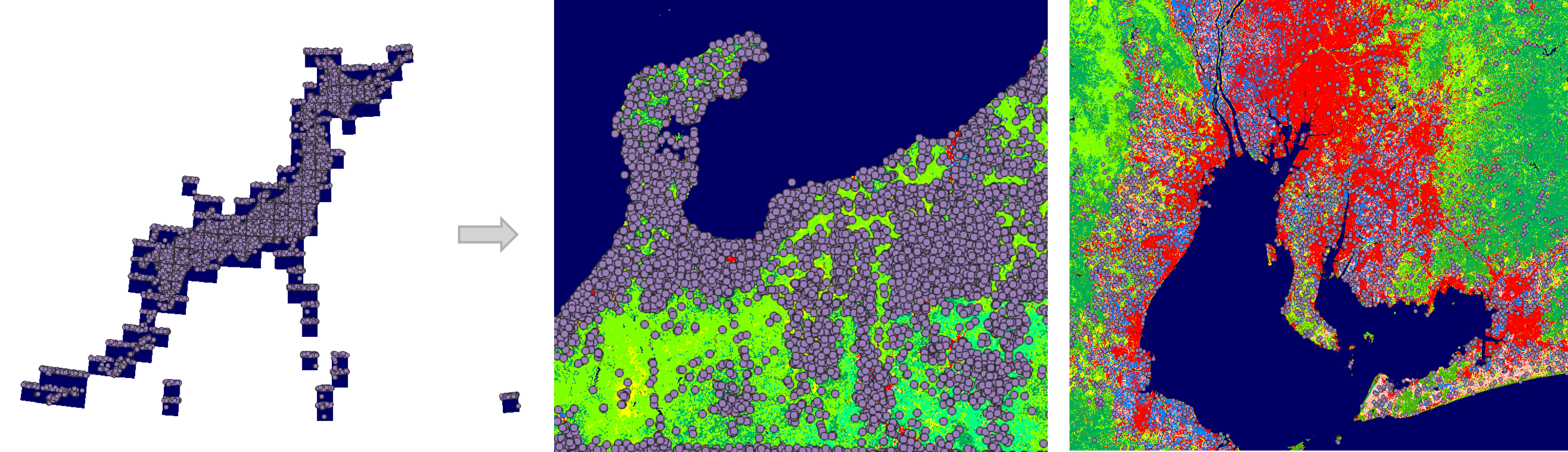}
  \caption{Category aware sampled locations on JAXA-LULC \cite{JAXA_LULC_2023} label data}
  \label{fig:patch_samples}
\end{figure}
\begin{figure}[!h]
  \centering
  \includegraphics[width=\textwidth]{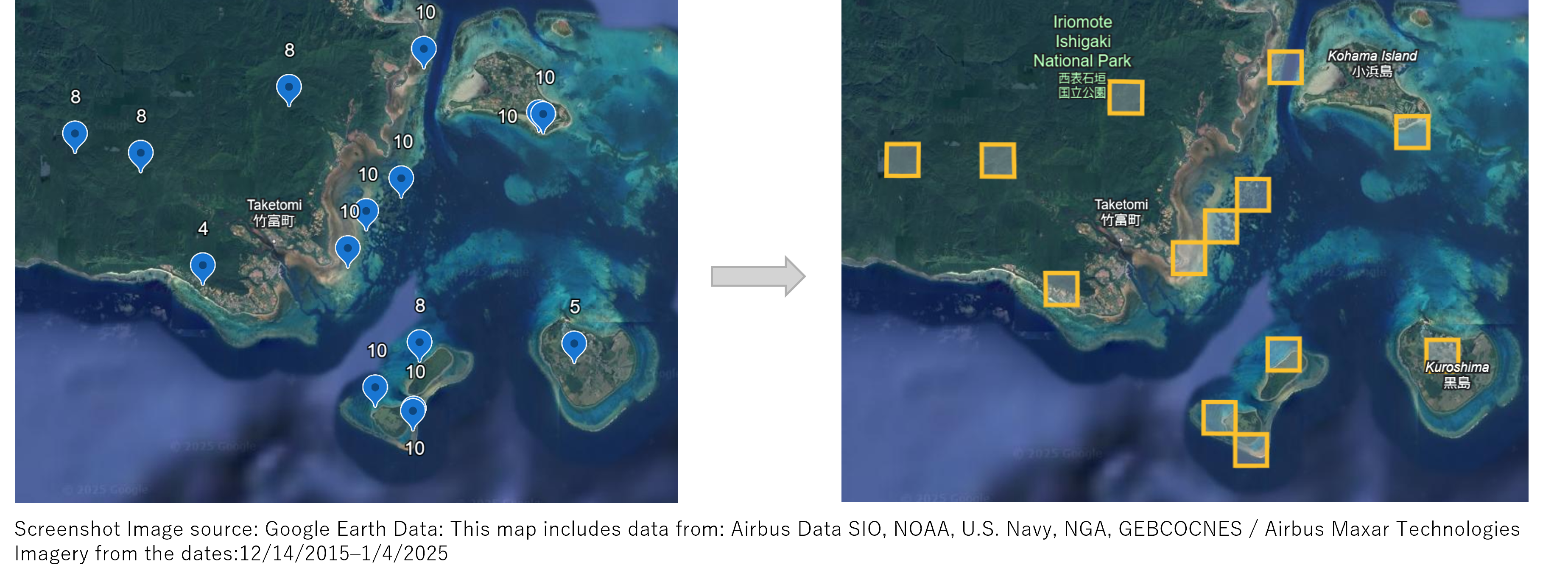}
  \caption{Sampled category-aware pixel locations and boundaries of corresponding data patches that consist of these sampled locations are represented on the map with their real world coordinates.}
  \label{fig:patch_samples}
\end{figure}

These sampled locations represents only one point corresponding to the sampled category. Next step would be to match these sampled points to the extracted SAR image patches comparing the latitude and longitude information. Due to huge number of sampled points within Japan, some of the sampled points may correspond to the same patch. So this will reduce number of patches that will be selected based on these sampled points. Moreover, due to large amount of forest area, we remove the sampled patches consisting of fully forest data.

Finally, based on these selections, we extracted more than 301,088 SAR images each size of 256x256 for pretraining. And, for segmentation fine-tuning task, we obtain more than 111,135 SAR images with the corresponding ground truth semantic labels. Fine-tuning data, then,  divided into three parts as: training: , validation, and test samples.

\section{Method}
\label{sec:method}

In our previous works \cite{sar_w_mixmae_igarss2025,sar_w_mixmae_ieeejstars2026,rssj2025sarwmixmae,caglayan2026enhancedlulcsegmentationlightweight}, to explore self-supervised pretraining on the SAR datasets, we took advantage of MixMAE \cite{liu2023mixmae}, a variant of masked auto-encoder that mixes two images for masked regions, and we extended it to proposed SAR-W-MixMAE (SAR weighted MixMAE) \cite{sar_w_mixmae_ieeejstars2026} by leveraging SAR-specific physical characteristics to apply intensity-based weighting to the auto-encoder training loss (mean absolute error). Then, we integrated UperNet \cite{xiao2018unified} segmentation decoder as in \cite{hong2024spectralgpt,rssj2025sarwmixmae,caglayan2026enhancedlulcsegmentationlightweight} to utilize downstream task for semantic segmantation of SAR images. 

In this study, we extend this concept to a more standard masked auto-encoder variant, SimMIM \cite{xie2022simmim} (where masked regions are zeroed), applied to ALOS-2 single-channel SAR imagery. Our approach, SAR-W-SimMIM, aims to mitigate the impact of speckle and extreme intensity values in SAR images. The primary objective is to evaluate the effect of self-supervised pretraining with SimMIM \cite{xie2022simmim} on semantic segmentation of SAR data and compare results against SAR-W-MixMAE and randomly initialized models. The SAR intensity-based weighting of the reconstruction loss has yielded encouraging results in both self-supervised SAR pretraining and downstream segmentation tasks. (Source code will be available at: https://github.com/nevrez/SAR-W-SimMIM).

\begin{figure}[!b]
  \centering
  \includegraphics[width=\textwidth]{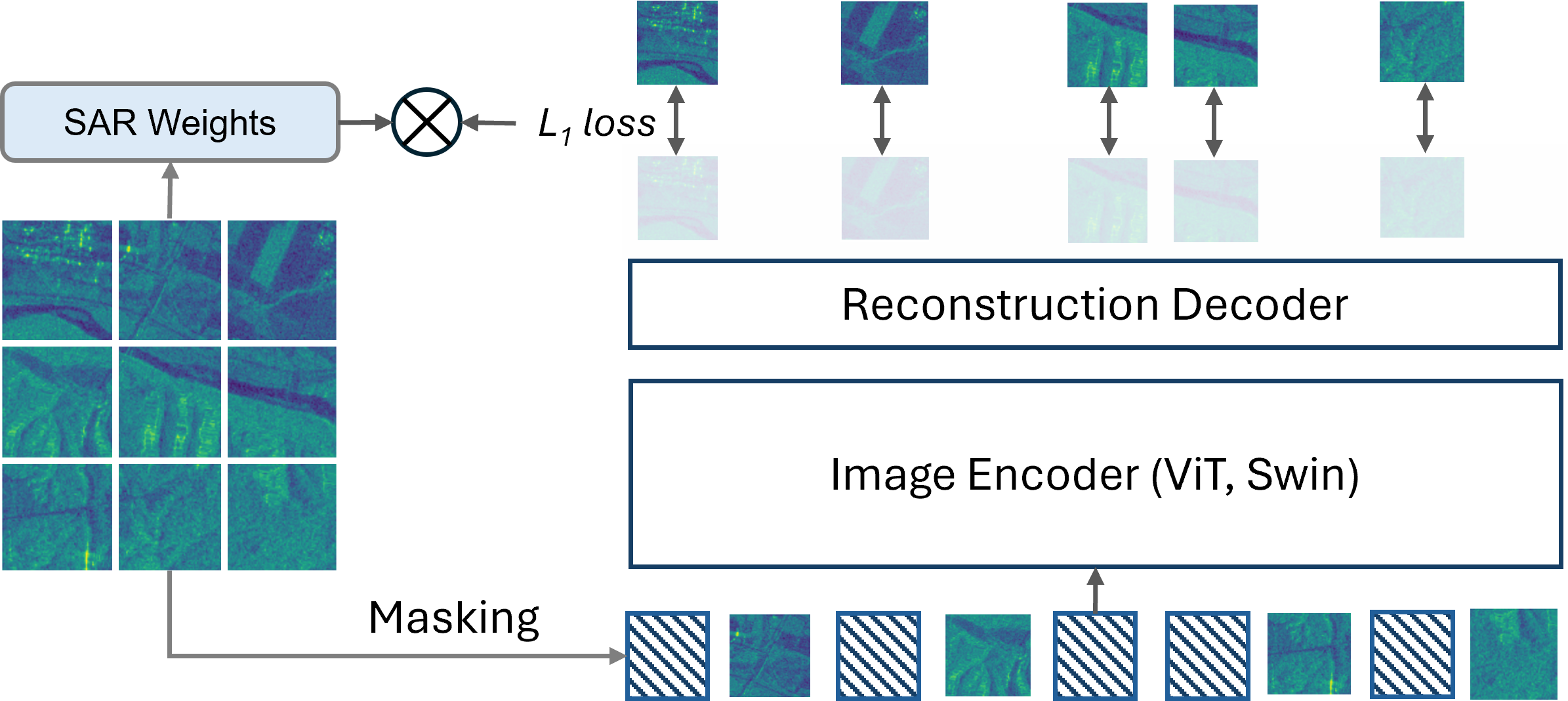}
  \caption{SAR-W-SimMIM: SAR-Weighted loss based self-supervised pretraining on SimMIM \cite{xie2022simmim}  }
  \label{fig:sar-w-simmim_model}
\end{figure}

\subsection{Self-Supervised Pretraining of SAR-Weighted-SimMIM}
\label{sec:method_pretraining}

\begin{figure}[!t]
  \centering
  \includegraphics[width=\textwidth]{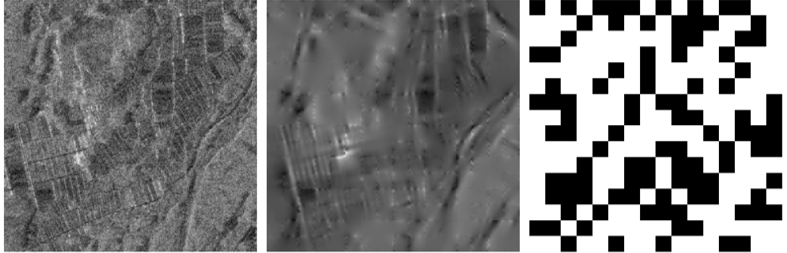}
  \caption{SAR-W-SimMIM reconstruction example}
  \label{fig:sar-w-simmim_reconstruction}
\end{figure}

 We trained our model on the original dataset without labels, using normalized input channel (HH) with SAR-W-SimMIM (see Figure \ref{fig:sar-w-simmim_model}). The inputs were normalized based on mean and standard deviation. The training of SAR-W-SimMIM is done by using the AdamW optimizer with a learning rate of $1 \times 10^{-4}$, which gradually decreased to $5 \times 10^{-7}$. Training lasted for 800 epochs, starting with a 40-epoch warm-up phase. In Figure \ref{fig:sar-w-simmim_reconstruction}, reconstruction sample is demonstrated after the pre-training is completed.

\subsection{SAR Image Segmentation Downstream Task}
\label{sec:method_finetuning}
\begin{figure}[!h]
  \centering
  \includegraphics[width=\textwidth]{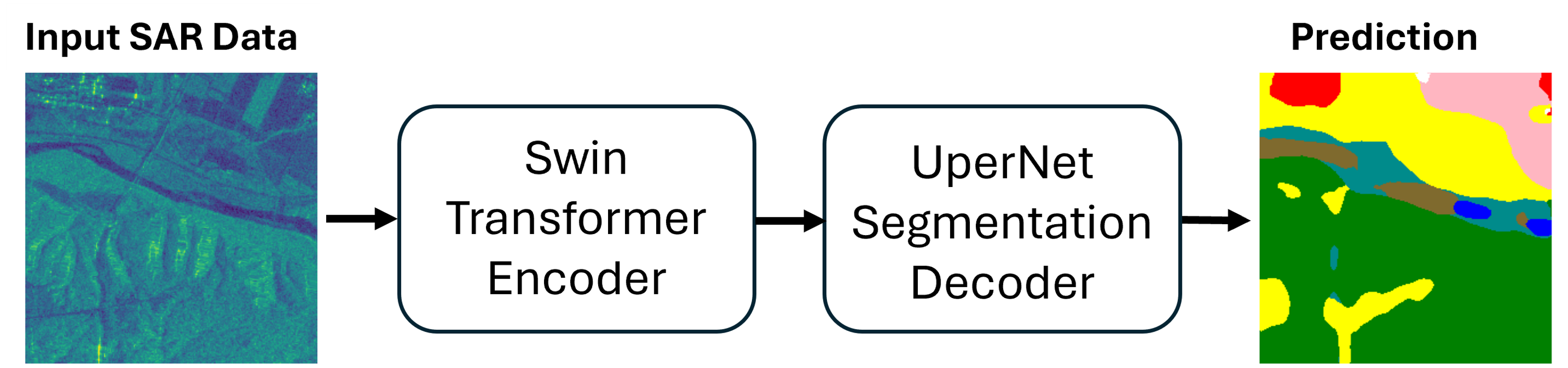}
  \caption{SAR Segmentation Task}
  \label{fig:encoder_upernet_segmentation}
\end{figure}

For fine-tuning, we replace the pretraining decoder with a UPerNet \cite{xiao2018unified} segmentation decoder (see Figure \ref{fig:encoder_upernet_segmentation}), following the approach in SpectralGPT \cite{hong2024spectralgpt}, with minor modifications to better integrate it with the SAR-W-MixMAE \cite{liu2023mixmae} encoder from the pretrained SAR-W-SIMMIM as similarly used in \cite{caglayan2026enhancedlulcsegmentationlightweight,rssj2025sarwmixmae}. Our task involves dense, pixel-wise semantic segmentation, requiring more extensive fine-tuning to effectively capture spatial context. The finetuning is done by using the AdamW optimizer with a learning rate of $1.25 \times 10^{-4}$, which gradually decreased to $2.5 \times 10^{-7}$. We fine-tuned the pretrained model for 100 epochs, including a 20-epoch warm-up phase.

For the loss function, we adopt a combination of focal loss and dice loss \cite{lin2017focal,milletari2016vnet} to address class imbalance and enhance segmentation accuracy.

~~~
~~~
The Dice Loss \cite{milletari2016vnet} is given by the formula:
\[
\text{Dice Loss} = 1 - \frac{2 \cdot \text{Intersection}}{\text{Union} + \epsilon}
\]
~~~where \(\text{Intersection}\) is the sum of the pixel-wise minimum of the predicted and ground truth values;  \(\text{Union}\) is the sum of the pixel-wise maximum of the predicted and ground truth values;  \(\epsilon\) is a small value to avoid division by zero. 

~~~
The Focal Loss \cite{lin2017focal} is defined as:
\[
\text{Focal Loss} = -\alpha \cdot (1 - p_t)^\gamma \cdot \log(p_t)
\]

~~~where \(p_t\) is the predicted probability for the true class;  \(\alpha\) is a balancing factor to address class imbalance; \(\gamma\) is the focusing parameter to adjust the rate at which easy examples are down-weighted.

The combined loss function is a weighted sum of Dice Loss and Focal Loss, where we balance the two losses with weight hyperparameters to have a balanced learning. We set Dice Loss weight to 0.32, Focal Loss weight to 0.57, \(\gamma = 1.1\) and \(\alpha = 0.35\) \cite{caglayan2026enhancedlulcsegmentationlightweight,rssj2025sarwmixmae}.

\section{Experimental Results}
\label{sec:experiments}

The fine-tuning experiments were conducted using the generated dataset, split into training (72,238 samples), validation (20,005 samples), and test (18,894 samples) sets. For this experiment, the five forest-related classes in the LULC dataset were merged into a single "forest" category, and a similar merging was applied to cropland-related classes. Moreover, any pacth with full forest covarege also removed from the data, which can also be done by setting sampling probability of forest to zero (see Section \ref{sec:category_sampling}). 

In Table \ref{tab:overall_comparison}, quantitative experiment results are given where MixMAE Baseline and SimMIM Baseline models mean that segmentation pretraining is done using their encoder together with the added UperNet segmentation decoder from scratch (random weights initialization). Fine-tuned results are based on the model encoders initially pre-training in a self-supervised manner. 

\begin{table}[!h]
    \centering
    \renewcommand{\arraystretch}{1.0}
    \setlength{\tabcolsep}{4pt}
    \begin{tabular}{lcc}
        \toprule
        Training Strategy & mAcc & mIoU \\
        \midrule        
        MixMAE Baseline (No Pretraining)    & 0.4223 & 0.3340 \\
        SAR-W-MixMAE (Fine-tuned)  & 0.5806 & 0.4734 \\
        SimMIM Baseline (No Pretraining)  & 0.5292 & 0.4258 \\
        SimMIM (Fine-tuned)  & 0.6064 & 0.4894 \\
        \textbf{SAR-W-SimMIM (Fine-tuned)}  & \textbf{0.6073} & \textbf{0.4898} \\
        \bottomrule
    \end{tabular}
    \caption{Overall accuracy (mAcc) and mean IoU (mIoU) comparison.}
    \label{tab:overall_comparison}
\end{table}

\begin{figure}[!h]
  \centering
  \includegraphics[width=\textwidth]{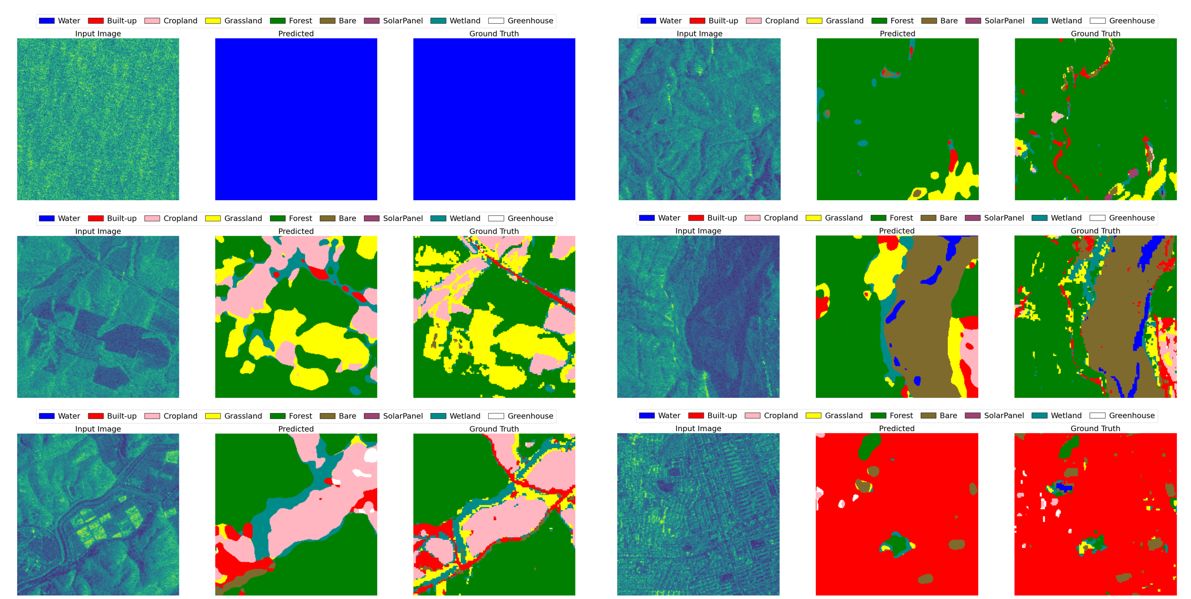}
  \caption{SAR Segmentation Results}
  \label{fig:patch_samples_segmentation_results}
\end{figure}

As shown in Table \ref{tab:overall_comparison}, fine-tuning with SAR-W-SimMIM model pretrained weights outperforms training from scratch and our previous trial with SAR-W-MixMAE for semantic segmentation downstream task, achieving higher mean accuracy and mean Intersection over Union (IoU) across categories. A qualitative example of inference over a large region is provided in Figure \ref{fig:patch_samples}, demonstrating strong visual agreement with the ground truth labels.

\section{Conclusion}
In this work, we explain how we construct a SAR dataset from ALOS2 single-channel imagery covering Japan, aimed at enabling efficient model pretraining and fine-tuning. The dataset is used to pretrain the SAR-W-SimMIM model, and the resulting encoder is fine-tuned for semantic segmentation using an added segmentation decoder. Experimental results on the segmentation task are promising, highlighting the potential for developing national-scale SAR foundation models in Japan.

\section{Acknowledgement}
{This work supported by AIST policy-based budget project ``R\&D on Generative AI Foundation Models for the Physical Domain".}{The ALOS-2 original data are copy-righted by JAXA and provided under the JAXA-AIST agreement.}{We used ABCI 3.0 provided by AIST and AIST Solutions with support from ``ABCI 3.0 Development Acceleration Use”.


%
%
\bibliographystyle{splncs04}
\bibliography{main}
\end{document}